%% file: HALIMI_Eusipco_2017_TR_v2.tex
\documentclass[12pt,technote,draftclsnofoot,onecolumn]{IEEEtran}

\usepackage{dsfont}
\usepackage{amsmath}
\usepackage{amssymb}
\usepackage{amsfonts}
\usepackage{graphicx}
\usepackage{amsmath}
\usepackage[all,poly]{xy}
\usepackage{multirow}
\usepackage{algorithm}
\usepackage{algorithmic}
\usepackage{color}
\usepackage[noadjust]{cite}
\usepackage{tikz,pgf}
\usetikzlibrary{fit}
\usetikzlibrary{arrows,automata}
\usepackage{verbatim}
\usetikzlibrary{positioning,shapes.geometric}
\usepackage{url}
\usepackage[T1]{fontenc}
\usepackage{subfig}

\newcommand{\figwidth}{\columnwidth}

\newcommand{\bmu}{\boldsymbol{\mu}}

\newcommand{\bthe}{\boldsymbol{\theta}}
\newcommand{\bsig}{\boldsymbol{\sigma}}

\input notations.tex

\title{AN UNSUPERVISED BAYESIAN APPROACH FOR THE JOINT RECONSTRUCTION AND
CLASSIFICATION OF CUTANEOUS REFLECTANCE CONFOCAL MICROSCOPY IMAGES}

\begin{document}

\author{Abdelghafour Halimi\thanks{This work was funded by Pierre Fabre Dermo Cosm\'{e}tique.}, Hadj Batatia,  Jimmy Le Digabel, Gwendal Josse and Jean-Yves Tourneret}

 \maketitle
\bigskip
\begin{center} \textbf{\textrm{TECHNICAL REPORT -- 2017, February}}\\
University of Toulouse, IRIT/INP-ENSEEIHT \\2 rue Camichel, BP 7122,
31071 Toulouse cedex 7, France
  \end{center}
\bigskip
\bigskip\bigskip

\begin{abstract} 
This paper studies a new Bayesian algorithm for the joint reconstruction
and classification of reflectance confocal microscopy
(RCM) images, with application to the identification of human
skin lentigo. The proposed Bayesian approach takes advantage of the distribution of the multiplicative speckle noise affecting the true reflectivity of these images and of appropriate priors for the unknown model parameters. A Markov chain Monte Carlo (MCMC) algorithm is proposed to jointly estimate
the model parameters and the image of true reflectivity
while classifying images according to the distribution of their
reflectivity. Precisely, a Metropolis-within-Gibbs sampler
is investigated to sample the posterior distribution of the Bayesian model associated with RCM images and to build estimators of its parameters, including labels indicating the class of each RCM image. The resulting algorithm is applied to synthetic data and to real images from a clinical study containing healthy and lentigo patients. 
\end{abstract}

\section{Introduction} \label{sec:Introduction} 
The lentigo is a hyperplasia that affects the skin. It comes from the proliferation of melanocyte cells at the dermo-epidermic junction. This leads to the disorganization of the regular cellular network \cite{menge2016concordance}. Clinically, this disorder is assessed visually on the skin surface or through biopsy. Reflectance confocal microscopy (RCM) imaging is increasingly used to explore various skin lesions \cite{Nehal2008,Hofmann2009}, including lentigo.  Figure 1 shows examples of images from patients with and without lentigo. Various studies have attested of the usefulness of RCM for cancer and other tumor diagnosis \cite{Alarconcar2014}. In \cite{menge2016concordance}, the authors reported good correlation between RCM and histology in the case of melanoma.  Studies of RCM for treatment follow-up \cite{Alarcon2014,Guitera2014,champin2014vivo} and guidance \cite{Hibler2015} have also been published.

\begin{figure*}[htbp]
\centering
\includegraphics[width=13cm, height=10cm]{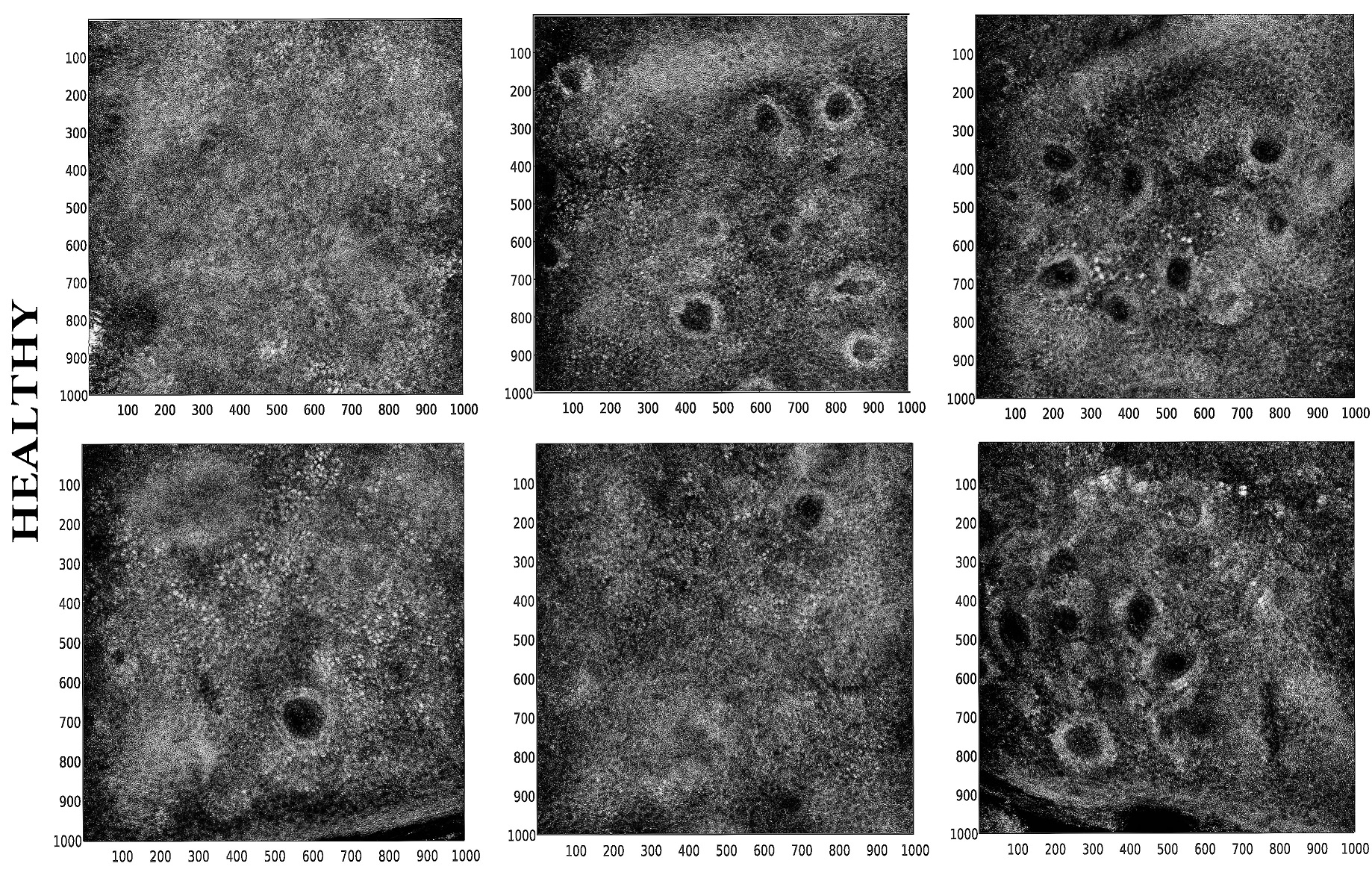}\\
\includegraphics[width=13cm, height=10cm]{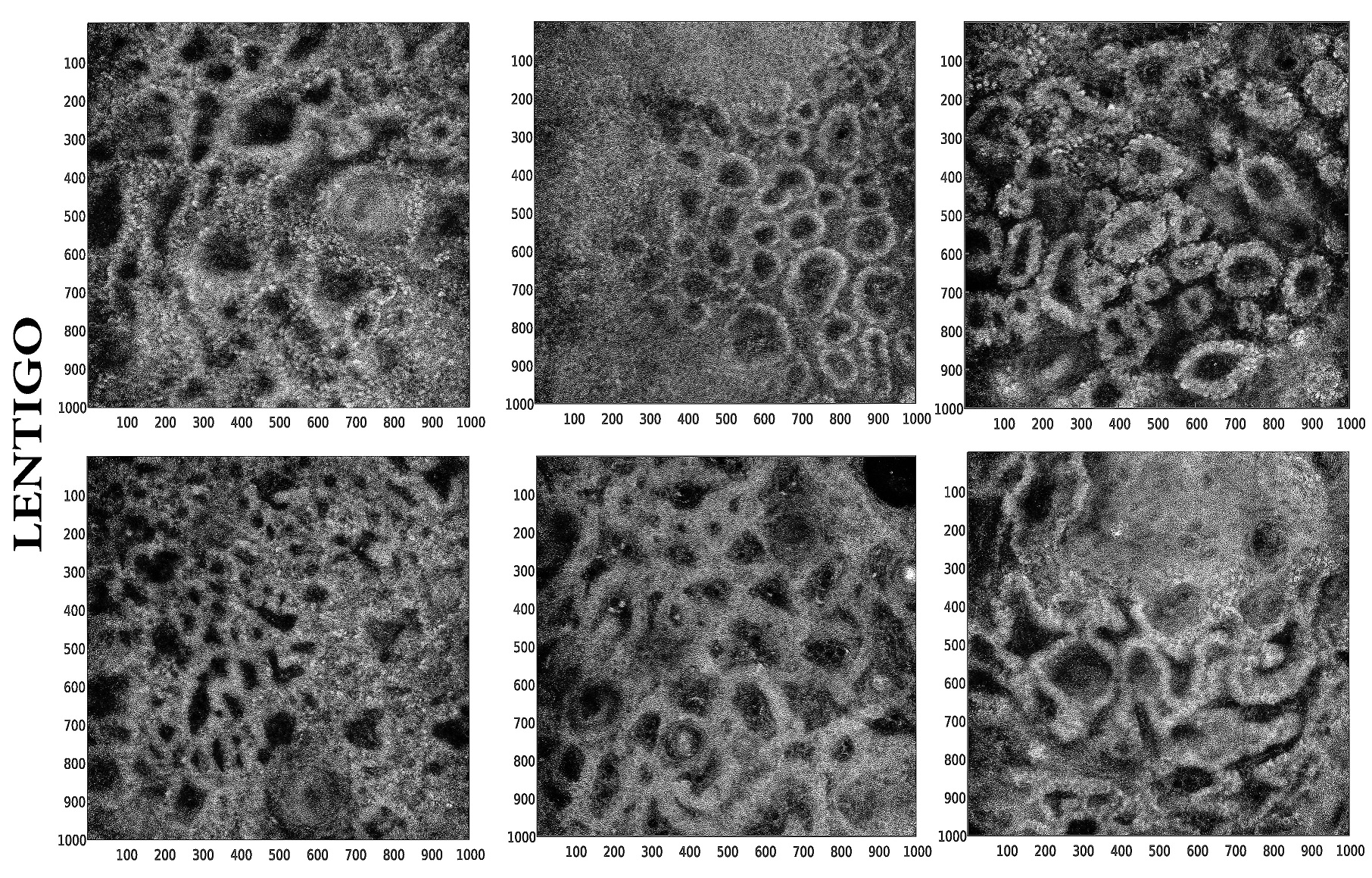}
\caption{\small Images (at the depth 49.5 $\mu m$) from healthy (patient $\#1$, $\#2$, $\#3$, $\#4$, $\#5$, $\#6$) and lentigo (patient $\#31$, $\#33$, $\#37$, $\#38$, $\#40$, $\#44$) patients at the DEJ depth. One can observe more textured images in the presence of lentigo.}
\label{fig:images1}
\end{figure*}


However, RCM images are up to now mainly analyzed visually. Image processing methods could be helpful to exploit their potential and provide aid for medical decision making.  Few of such methods were reported in the literature. In \cite{Luck2005}, Luck et al. developed a nuclei segmentation method based on a Gaussian model for the nuclei reflectivity and a truncated Gaussian distribution for the intensity of the cytoplasm fibers. Their bayesian classification algorithm relies on a Gaussian Markov random field to ensure spatial correlation. Another application of RCM was developed and validated by Kurugol et al. to identify the dermoepidermal junction. This method is based on statistical classification of texture features \cite{kurugol2011semi,kurugol2012validation}. Hames et al. \cite{hames2015anatomical,hames2016automated} proposed a skin layer segmentation method based on a logistic regression classifier. An SVM classification method was developed in \cite{kose2016machine} based on speeded up robust features. This method was applied to identifying skin morphological patterns using RCM image texture. Finally, a wavelet-based classification method was developed in \cite{koller2011vivo} to distinguish benign and malignant melanocytic skin tumors. This method, with which we compare our's, consists of classifying a vector of 39 features using a decision tree approach. In this paper we proposed a bayesian method to jointly reconstruct RCM true reflectivity images while classifying images as lentigo or healthy.

The first contribution of this paper is a hierarchical Bayesian model that allows a set of RCM images to be classified into healthy and lentigo classes. Each image is assumed to be corrupted by a multiplicative speckle noise with a gamma distribution. A truncated Gaussian distribution is then assigned to each image to classify, constraining these images to be positive. Prior distributions are finally assigned to the means and variances of these truncated Gaussian distributions, to the noise variances, and to the image labels. The joint posterior distribution of the proposed model is finally determined and will be
used for image classification and parameter estimation. The second contribution of this paper is the derivation of an estimation algorithm associated with the proposed hierarchical Bayesian model. As the minimum mean square error (MMSE) and maximum a posteriori (MAP) estimators of this model cannot be easily computed from its joint posterior, we investigate a hybrid Gibbs sampler allowing the posterior of interest to be samples (see \cite{chib1995understanding,gilks1995markov} for details). The proposed Bayesian model and estimation algorithm are validated using synthetic and real RCM images, resulting from a clinical study containing healthy and lentigo patients.  The obtained results are very promising and show the potential of the proposed denoising and classification strategy. 

The paper is structured as follows. The classification problem studied in this work is introduced in Section
\ref{sec:Problem_formulation}. The proposed hierarchical Bayesian model and its estimation algorithm are studied in Sections \ref{sec:Hierarchical_Bayesian_model} and \ref{sec:METROPOLIS}. Section \ref{subsec:Simulation_results_on_synthetic_data} validates the proposed technique using simulated data with different noise levels. Section \ref{sec:Results_on_Jason2_real_data} shows results obtained using real data obtained from a clinical study. Conclusions and future work are finally reported in Section
\ref{sec:Conclusions}.

\section{Problem formulation} \label{sec:Problem_formulation}

\subsection{Observation model} 
Consider $L$ noise free images, containing N pixels,  gathered in the matrix $\bsS=\left[\bss_1, \cdots,  \bss_l \right] \in \mathds{R}^{N\times L}$, where $\bss_l$, $l\in \{1,\cdots,L\}$ denotes the image associated with the $l$th patient. Denote by $\bsY=\left[\bsy_1, \dots,  \bsy_l \right] \in \mathds{R}^{N\times L}$  the corresponding noisy images.     
Using these notations, the observation model is given by  
\begin{equation}
\bsy_l = \bss_l \odot \bsb_l, \textrm{  with   } \bsb_l \sim \mathcal{G}(\rho_l,\theta_l)
\label{eqt:Observation_model}
\end{equation}
where $\bsy_l$ and $\bss_l$ are  ($N \times 1$) vectors representing the $l$th observed and noiseless images,  $\bsb_l$ is a gamma noise ($N \times 1$) vector with a shape parameter $\rho_l$ and a scale parameter $\theta_l$ and $\odot$ denotes the termwise product. In order to ensure that the proposed model \eqref{eqt:Observation_model}  is identifiable, the mean of the gamma noise is supposed to equal $1$, leading to
\begin{equation}
 \mathbb{E} (\bsb_{l}) =1 \ \ \ \Rightarrow  \ \ \rho_l=\frac{1}{\theta_l}.
\label{eqt:noise}
\end{equation}
The problem addressed in this paper is to classify these images $\bsy_l$, $l\in \{1,\cdots,L\}$  into two classes representing healthy  and lentigo patients. The next section introduces a hierarchical Bayesian model that is used for this classification.


\section{Hierarchical Bayesian model} \label{sec:Hierarchical_Bayesian_model}
This section introduces a hierarchical Bayesian model that can be used to estimate the unknown $N\times L$ matrix of noiseless images $\bsS$, the $L \times 1$ vectors $(\bsz,\boldsymbol{\theta})$ containing the class labels and the noise variances associated with the $L$ observed images from the matrix $\bsY$. This model is defined by a likelihood, and by parameter and hyperparameter priors defined below.

\subsection{Likelihood} \label{subsec:Likelihood}
The multiplicative speckle noise $\bsb_l$ is known to have a gamma distribution. Thus, the observation model \eqref{eqt:Observation_model} leads to
\begin{equation}
 y_{nl}| s_{nl},\theta_{l}\sim \mathcal{G}\left(\frac{1}{\theta_l},s_{nl}\ \theta_l\right)\ \ \  
\label{eqt:likelihood2}
\end{equation}
where $\sim$ means "is distributed according to", $\mathcal{G}$ is the gamma distribution whose probability density function (pdf) is
\begin{equation}
f(y_{nl}\mid s_{nl},\theta_l)\propto \frac{\left(y_{nl}\right)^{\frac{1}{\theta_l}-1}  \exp\left(-\frac{y_{nl}}{s_{nl}\ \theta_l}\right)}{\Gamma\left(\frac{1}{\theta_l}\right) 
\left(s_{nl} \ \theta_l\right)^\frac{1}{\theta_l}}  I_{\mathbb{R}^+}(y_{nl}) \label{eqt:likelihood}
\end{equation}
with $I_{\mathbb{R}^+}(y_{nl})$ the indicator function on $\mathbb{R}^+$, $\propto$ means ``proportional to'' and $ \Gamma $ denotes the gamma function. Assuming independence between the observed signals, the likelihood of the $L$ observed images can be written
$$
f(\bsY| \bsS,\boldsymbol{\theta}
 ) \propto \prod_{n=1}^N\prod_{l=1}^{L}{f(y_{nl}|s_{nl}, \theta_l)}.
 $$

\subsection{Priors for the signal of interest} \label{subsec:Priors_for_the_observed_signal} 
To ensure the positivity of the noiseless images, a truncated Gaussian distribution is assigned to $\bss_l$ for $l\in \{1,\cdots,L \}$ 
\begin{equation}
\bss_l\mid z_{l}=k, \mu_k,\sigma_k^2  \sim \mathcal{N}_{\mathds{R^+}}(\mu_k,\sigma_k^2)  \ 
\label{eqt:priorD}
\end{equation}
where $\mathcal{N}_{\mathcal{S}}$ denotes the truncated normal distribution on $\mathcal{S}$, $k$ takes the two values $1$ and $2$ depending on the patient class,  and $(\mu_k,\sigma_k^2)$ are the means and variance of the two truncated Gaussian distributions. 

\subsection{Prior for the noise variances} \label{subsec:Prior_for_the_Noise_variance} 
A non-informative conjugate inverse gamma prior (denoted as $ \mathcal{IG}$) is classically selected for the scale parameter $\theta_j$ \cite{fink1997compendium} 
\begin{equation}
\theta_{l}\mid a,b \sim \mathcal{IG}(a,b)
\end{equation}
where $a$ and $b$ are fixed hyperparameters, that are adjusted to reflect the absence of prior knowledge on $\theta_l$, i.e., the mean and variance of $\theta_{l}$ were fixed to $1$ and $100$ in order to obtain a flat prior. The joint prior for the vector of noise variances denoted as $f(\boldsymbol{\theta}\mid a,b)$ is finally obtained as the product of the marginal densities $f(\theta_i \mid a,b)$.

\subsection{Prior for the label vector $\bsz$} \label{subsec:Prior_for_the_label} 
The parameter vector $\bsz=(z_1,...,z_L)$ is a label vector that associates each image to a healthy or lentigo skin. Because of the absence of prior knowledge about this parameter, it is assigned a uniform prior defined as  
\begin{equation} \label{prior_z}
P(z_{l}=k)=\frac{1}{2}, \forall l=1,...,L.
\end{equation} 
The labels associated with the different patients are supposed to be a priori independent, i.e., the joint prior of $\bsz$ denoted as f($\bsz$) is the product of the probabilities defined in \eqref{prior_z}.

\subsection{Hyperparameter priors} \label{subsec:Hyperparameter_priors}
In order to complete the description of the proposed hierarchical Bayesian model and to allow hyperparameters to be estimated directly from the data, we propose to assign priors for the different hyperparameters. A Gaussian prior has been selected for the mean $\mu_{ k}$ and a non-informative  inverse gamma prior for the variance $\sigma_{ k}^2$ (see \cite{Dobigeon2008,fink1997compendium} for motivations)
\begin{equation}
\mu_{k}\mid \mu_0,\sigma_0 \sim \mathcal{N}(\mu_0,\sigma_0^2) \ \ \  
\end{equation}
\begin{equation}
\sigma_{k}^2\mid \alpha_0,\beta_0 \sim \mathcal{IG}(\alpha_0,\beta_0)
\end{equation}
where $\mu_0$, $\sigma_0^2$,  $\alpha_0$, $\beta_0$ are fixed in order to obtain flat priors, i.e., $\mu_0 = 100$, $\sigma_0^2= 10^5$ whereas the mean and variance of $\sigma_{k}^2$ were fixed to $1$ and $1000$. The joint pdfs $f(\bmu\mid \mu_0,\sigma_0)$ and $f(\bsig^2\mid \alpha_0,\beta_0)$ are finally obtained as the product of their marginal densities assuming prior independency between the components of these two vectors.

\subsection{Joint posterior distribution} \label{subsec:Posterior_distributions}
\begin{figure}
\centering
\begin{tikzpicture}
 nodes %
\node[text centered] (Y) {$\bsY$};
\node[above =1. of Y, text centered] (top1) {$ $};
\node[above =1. of top1, text centered] (top2) {$ $};
\node[above =1. of top2, text centered] (top3) {$ $};
\node[right  =1  of top1, text centered] (THE) {$\bthe$};
\node[left  =1.2  of top1, text centered] (S) {$\bsS$};
\node[draw, rectangle, right =1.7  of top2, text centered] (b) {$b$};
\node[draw, rectangle, right =0.5  of top2, text centered] (a) {$a$};

\node[left  =0.  of top2, text centered] (SIG) {$\bsig_k$};
\node[left  =2.1  of top2, text centered] (MU) {$\bmu_k$};
\node[left  =3.5 of top2, text centered] (Z) {$\bsz$};

\node[draw, rectangle, left  =2.8  of top3, text centered] (MU0) {$\mu_0$};
\node[draw, rectangle, left  =1.6  of top3, text centered] (SIG0) {$\sigma_0$};
\node[draw, rectangle, left  =0.6  of top3, text centered] (ALPHA0) {$\alpha_0$};
\node[draw, rectangle, right  =0.  of top3, text centered] (BETA0) {$\beta_0$};

\node[draw,dashed,fit=(MU) (SIG)] {};
edges %
\draw[->, line width= 1] (S) to  [out=270,in=155, looseness=0.5] (Y);
\draw[->, line width= 1] (THE) to  [out=270,in=35, looseness=0.5] (Y); 
\draw[->, line width= 1] (b) to  [out=270,in=35, looseness=0.5] (THE);
\draw[->, line width= 1] (a) to  [out=270,in=145, looseness=0.5] (THE);
\draw[->, line width= 1] (SIG) to  [out=270,in=35, looseness=0.5] (S);
\draw[->, line width= 1] (MU) to  [out=270,in=145, looseness=0.5] (S);
\draw[->, line width= 1] (Z) to  [out=270,in=165, looseness=0.5] (S);

\draw[->, line width= 1] (SIG0) to  [out=270,in=35, looseness=0.5] (MU);
\draw[->, line width= 1] (MU0) to  [out=270,in=145, looseness=0.5] (MU);

\draw[->, line width= 1] (ALPHA0) to  [out=270,in=145, looseness=0.5] (SIG);
\draw[->, line width= 1] (BETA0) to  [out=270,in=35, looseness=0.5] (SIG);

\end{tikzpicture}
 
\caption{DAG for the parameter and hyperparameter priors.
The user fixed hyperparameters appear in  boxes (continuous line). }
\label{fig:dag}
\end{figure}
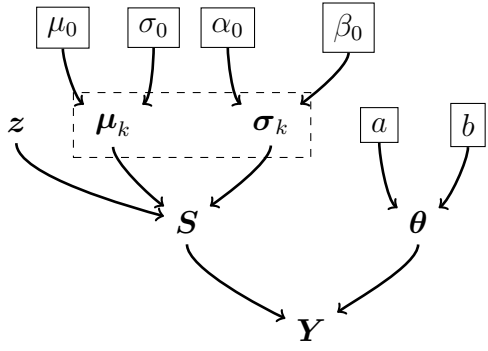

The proposed Bayesian model is illustrated by the directed
acyclic graph (DAG) displayed in Fig. \ref{fig:dag}, which highlights the
relationships between the observations $\bsY$, the parameters $\bsS,  \boldsymbol{\theta}, \bsz$ and the hyperparameters  $\mu_k,\sigma_k^2$. Assuming prior independence
between the different components of the parameter vector $\bsX = \left(\bsS,\boldsymbol{\theta},\bsz,\mu_k,\sigma_k^2 \right)$, the joint posterior distribution of this Bayesian model can be computed using the following hierarchical structure 
\begin{equation}
f(\bsX\mid\bsY)\propto f(\bsY\mid\bsS,\boldsymbol{\theta} ) f(\bsS,\boldsymbol{\theta},\bsz,\bmu,\bsig^2)
\label{eqt:posterior}
\end{equation}
\begin{equation}
\text{with}\ \ f(\bsS,\boldsymbol{\theta},\bsz,\bmu,\bsig^2) = f(\bsS\mid \bsz, \bmu,\bsig^2) f(\boldsymbol{\theta}\mid a,b)   \  f(\bmu\mid \mu_0,\sigma_0)f(\bsig^2\mid \alpha_0,\beta_0) f(\bsz).
\end{equation}

The complexity of the proposed Bayesian model summarized in the DAG of Fig. \ref{fig:dag} and its resulting posterior \eqref{eqt:posterior} prevent a simple computation of the maximum a-posteriori (MAP) or
minimum mean square (MSE) estimators of the unknown model parameters. The next section studies an MCMC method that is used to sample the posterior \eqref{eqt:posterior} and to build estimators of the parameters involved in the proposed Bayesian model using the generated samples.

\section{Metropolis-within-Gibbs algorithm} \label{sec:METROPOLIS}
This section studies a hybrid-Gibbs-sampler, which is guaranteed to generate samples asymptotically distributed according to the target distribution \eqref{eqt:posterior}. The Gibbs sampler described in Algo. \ref{alg:Metropolis}, iteratively generates samples distributed according to the conditional distributions of \eqref{eqt:posterior}. These conditional distributions are detailed in the rest of this section. Because of the complexity of the conditional distributions, we consider random-walk Metropolis-Hastings (RWMH) \cite{gilks1995markov,chib1995understanding} moves within the Gibbs sampler, which consists in generating samples distributed according to the complex
conditional distribution of each parameter of interest. This is achieved using the conditional distributions $f_j(.)$, for $j \in \{1,...,J \}$, and their associated proposal distributions $g_j(.)$. The first step is to initialize the sample value for each parameter $\bsX_j^{(0)}$, for $j\in \{1,...,J \}$. The main loop of the RWMH algorithm consists of three components: 
\begin{enumerate}
\item Generate a candidate $\bsX_j^{\textrm{cand}}$ from the proposal
distribution 
$g_j\left(\bsX_{j}^{(\text{cand})}\mid\bsX_{j}^{(i-1)}\right)$
. The distribution $g_j(.)$ is the truncated Gaussian distribution \cite{robert1995simulation} $\mathcal{N}_{\mathds{R^+}}\left(\bsX_{j}^{(i-1)},\bf\epsilon_j^2\right)$  for the parameters $\bsS,\boldsymbol{\theta},\bsig^2,$ and the Gaussian distribution $\mathcal{N}\left(\bsX_{\bmu}^{(i-1)},\bf\epsilon_{\bmu}^2\right)$ for $\bmu$.
\item Compute the acceptance probability using the acceptance function $\alpha\left(\bsX_{j}^{(\textrm{cand})}\mid\bsX_{j}^{(i-1)}\right)$ based upon the proposal distribution and the conditional density for each parameter  \\  $ \alpha\left(\bsX_{j}^{(\textrm{cand})}\mid\bsX_{j}^{(i-1)}\right)= \min \left\{ \frac{f_j\left(\bsX_{j}^{(\textrm{cand})}\right)}{f_j\left(\bsX_{j}^{(i-1)}\right)} \ \  \frac{g_j\left(\bsX_{j}^{(i-1)}\mid\bsX_{j}^{(\textrm{cand})}\right)}{g_j\left(\bsX_{j}^{(\textrm{cand})}\mid \bsX_{j}^{(i-1)}\right)} \ ,1\right\}$
\item Accept the candidate with probability $\alpha\left(\bsX_{j}^{(\textrm{cand})}\mid\bsX_{j}^{(i-1)}\right)$. 

\end{enumerate}

In order to maximize the efficiency of the algorithm, the variances $\bf\epsilon_j^2$ of the proposal distributions    have been adjusted such that the acceptance rate is between 0.3 and 0.6 as suggested in  \cite{chib1995understanding} and \cite{roberts2001optimal}.

\begin{algorithm}
\caption{Metropolis-within-Gibbs algorithm} \label{alg:Metropolis}
\begin{algorithmic}[1]
       \STATE Input: $\textrm{N}_{\textrm{bi}},\textrm{N}_{\textrm{MC}}, \bsS,\boldsymbol{\theta},\bsz,\bmu,\bsig^2 $
       \STATE \underline{Initialization}
       \STATE Initialize $\bsS^{(0)}, \boldsymbol{\theta}^{(0)}, \bsz^{(0)}, \bmu^{(0)}, \bsig^{2(0)}$     
       \FOR{$i$=1 to $N_{\textrm{MC}}$}
       \STATE \underline{Parameter update}
       \STATE Sample $\bsS^{(i)} \mid \bsY,  \boldsymbol{\theta}, \bsz, \bmu, \bsig^{2}$ according to \eqref{eqt:posteriorS}  using an RWMH with a truncated Gaussian proposal    
       \STATE Sample $\boldsymbol{\theta}^{(i)} \mid \bsY, \bsS, a, b$ according to \eqref{eqt:posteriortheta}  using an RWMH with a truncated Gaussian proposal    
       \STATE Sample $\bmu^{(i)} \mid \bsS,\bsig^2, \mu_0,\sigma_0^2$ according to \eqref{eqt:posteriormu}  using an RWMH with a  Gaussian proposal    
       \STATE Sample $\bsig^{2(i)} \mid \bsS, \bmu, \alpha_0, \beta_0$ according to \eqref{eqt:posteriorsigma}  using an RWMH with a truncated Gaussian proposal    
      \STATE Sample $\bsz^{(i)} \mid \bsS, \bmu, \bsig^2$ from the pdf \eqref{eqt:posteriorZ}  
       \ENDFOR
                  
\STATE Result: $\bsS^{(i)},\boldsymbol{\theta}^{(i)}, \bsz^{(i)}, \bmu^{(i)},\bsig^{2(i)} $ for $i=1,...,\textrm{N}_{\textrm{MC}}$.			
\end{algorithmic}
\end{algorithm}

The conditional distributions of the parameters of interest
are obtained by multiplying the likelihood with the different priors and by removing the multiplicative terms that do not depend on the variable of interest. The algorithm iteratively updates each parameter by using
its conditional distribution detailed in the following paragraphs.

\subsection{Sampling the parameter $s_{n l}$ :} 
\begin{equation}
\small f\left(s_{n l}\mid z_{l}=k,y_{n l},\theta_{l},\sigma_k^2,\mu_{ k} \right)\propto \frac{1}{\left(s_{n l}\right)^{1/\theta}}  \ \ \exp\left(-\frac{y_{nl}}{s_{nl}\ \theta_l}\right) \\ \times  \  \exp\left[-\frac{1}{2\ \sigma_k^2 }\left(s_{nl}-\mu_k\right)^2   \right] I_{\mathbb{R}^+}(s_{nl})
\label{eqt:posteriorS}
\end{equation}

\text{The proposal law is :}
\begin{center}
\begin{equation*}
g\left(x\mid s_{nl}^t\right)   \propto \  \ \frac{\frac{1}{\sqrt{2\ \pi}\ \epsilon_2} \ \exp \left[- \frac{(x-s_{nl}^t)^2}{2\ \epsilon_2^2}\right]}{1-\Phi\left(\frac{-s_{nl}^t}{\epsilon_2}\right)}. \ \ \ 
\end{equation*}
\end{center}

We generate $s_{nl}^*$, such that : 
\begin{equation*}
 s_{nl}^* \sim g\left(x\mid s_{nl}^t\right).
\end{equation*}
$\textrm{The acceptance-rejection rule is :}$ \\
\begin{align*}
 \ \ \ s_{nl}^{t+1}=\begin{cases}s_{nl}^* \ \ \ \textrm{with \ \ prob} \ \  \min \left\{ \frac{f(s_{nl}^*)}{f(s_{nl}^t)} \ \  \frac{g(s_{nl}^t\mid s_{nl}^*)}{g(s_{nl}^*\mid s_{nl}^t)} \ ,1\right\}
\\  s_{nl}^t \ \  \ else\end{cases}
\end{align*}

with 
\begin{equation*}
 \frac{f(s_{nl}^*)}{f(s_{nl}^t)} \ \  \frac{g(s_{nl}^t\mid s_{nl}^*)}{g(s_{nl}^*\mid s_{nl}^t)} =  \left(\frac{s_{nl}^*}{s_{nl}^t}\right)^{-\rho_l} \ \ \left[\frac{1-\Phi\left(\frac{-s_{nl}^t}{\epsilon_2}\right)}{1-\Phi\left(\frac{-s_{nl}^*}{\epsilon_2}\right)}\right] \times
\end{equation*}
\begin{small}
\begin{equation*}
  \exp\left[\frac{-2 \ y_{nl}\ \sigma_k^2 - s_{nl}^*\ \theta_l \ (s_{nl}^*-\mu_k)^2 }{2 \ \sigma_k^2 \ s_{nl}^* \ \theta_l}+\frac{2 \ y_{nl}\ \sigma_k^2 +s_{nl}^t\ \theta_l \ (s_{nl}^t-\mu_k)^2 }{2 \ \sigma_k^2 \ s_{nl}^t \ \theta_l}\right]. \ \ 
\end{equation*}
\end{small}
\\
\subsection{Sampling the parameter $\theta_{l}$ :} 

\begin{equation}
\small f\left(\theta_{l}\mid y_{nl},s_{nl},a,b\right) \propto \frac{1}{\theta_l^{\ N/\theta_l+a+1}} \ \prod_{n=1}^N\left( \frac{y_{nl}}{s_{nl}}\right)^{\frac{1}{\theta_l}}\ \ \\ \times \exp\left[   -\frac{1}{\theta_l} \left(\sum_{n=1}^N\frac{y_{nl}}{s_{nl}}-b\right)\right] \  \ \left[\Gamma\left(1/\theta_l\right)\right]^{-N}
\label{eqt:posteriortheta}
\end{equation}

\text{The proposal law is :}
\begin{center}
\begin{equation*}
g\left(x\mid \theta_{l}^t\right)   \propto \  \ \frac{\frac{1}{\sqrt{2\ \pi}\ \epsilon_1} \ \exp \left[- \frac{(x-\theta_{l}^t)^2}{2\ \epsilon_1^2}\right]}{1-\Phi\left(\frac{-\theta_{l}^t}{\epsilon_1}\right)}. \ \ \ 
\end{equation*}
\end{center}
$\text{where $\Phi$ is the cumulative normal distribution function}$

We generate $\theta_{l}^*$, such that : 
\begin{equation*}
 \ \theta_{l}^* \sim g\left(x\mid \theta_{l}^t\right).
\end{equation*}
$\textrm{The acceptance-rejection rule is :}$ \\
\begin{equation*}
 \ \ \ \ \theta_{l}^{t+1}=\begin{cases}\theta_{l}^* \ \ \ \textrm{with \ \ prob}\ \  \min \left\{ \frac{f(\theta_{l}^*)}{f(\theta_{l}^t)} \ \  \frac{g(\theta_{l}^t\mid \theta_{l}^*)}{g(\theta_{l}^*\mid \theta_{l}^t)} \ ,1\right\}
\\  \theta_{l}^t \ \  \ else\end{cases}
\end{equation*}

with 
\begin{equation*}
 \frac{f(\theta_{l}^*)}{f(\theta_{l}^t)} \ \  \frac{g(\theta_{l}^t\mid \theta_{l}^*)}{g(\theta_{l}^*\mid \theta_{l}^t)} =  \frac{{\theta_{l}^*}^{\left(-\frac{N}{\theta_{l}^*}-a-1\right)}}{{\theta_l^t}^{\left(-\frac{N}{\theta_l^t}-a-1\right)}} \ \left[\frac{\Gamma\left(\frac{1}{\theta_l^t}\right)
}{\Gamma\left(\frac{1}{\theta_{l}^*}\right)
}\right]^N  \times \ \exp\left[\frac{-\left(\sum_{n=1}^N\frac{y_{nl}}{s_{nl}}\right)-b}{\theta_{l}^*}+\frac{\left(\sum_{n=1}^N\frac{y_{nl}}{s_{nl}}\right)+b}{\theta_l^t}\right] \ 
\end{equation*}
\begin{equation*}
\times \ \frac{\left(\prod_{n=1}^N \frac{1}{s_{nl}}\right)^\frac{1}{\theta_{l}^*} \  \left(\prod_{n=1}^N y_{nl}\right)^{\frac{1}{\theta_{l}^*}-1}}{\left(\prod_{n=1}^N \frac{1}{s_{nl}}\right)^\frac{1}{\theta_l^t} \  \left(\prod_{n=1}^N y_{nl}\right)^{\frac{1}{\theta_l^t}-1}}   \  \left[\frac{1-\Phi\left(\frac{-\theta_l^t}{\epsilon_1}\right)}{1-\Phi\left(\frac{-\theta_{l}^*}{\epsilon_1}\right)}\right].
\end{equation*}
\\

\subsection{Sampling the parameter $\mu_k$ :}

\begin{equation}
f\left(\mu_k\mid s_{nl},\sigma_k^2,\mu_0,\sigma_0\right) \propto\frac{\exp \left[ \frac{-\sum_{n=1}^N\sum_{l=1}^{L_k} \ (s_{nl}-\mu_k)^2}{2\ \sigma_k^2}-\frac{(\mu_k-\mu_0)^2}{2\ \sigma_0^2}\right]}{\left(1-\Phi\left(-\frac{\mu_k}{\sigma_k}\right)\right)^{NL_k}}
\label{eqt:posteriormu}
\end{equation}

\text{The proposal law is :}
\begin{center}
\begin{equation*}
g\left(x\mid \mu_k^t\right)   \propto \  \ {\frac{1}{\sqrt{2\ \pi}\ \epsilon_3} \ \exp \left[- \frac{(x-\mu_k^t)^2}{2\ \epsilon_3^2}\right]}.\ \ \ 
\end{equation*}
\end{center}

We generate $\mu_k^*$, such that : 
\begin{equation*}
  \mu_k^* \sim g\left(x\mid \mu_k^t\right).
\end{equation*}
$\textrm{The acceptance-rejection rule is :}$ \\
\begin{equation*}
 \ \ \ \ \mu_k^{t+1}=\begin{cases}\mu_k^* \ \ \textrm{with \ \ prob} \  \min \left\{ \frac{f(\mu_k^*)}{f(\mu_k^t)} \ \  \frac{g(\mu_k^t\mid \mu_k^*)}{g(\mu_k^*\mid \mu_k^t)} \ ,1\right\}
\\  \mu_k^t \ \  \ else\end{cases}
\end{equation*}
\\
with 

\begin{equation*}
 \frac{f(\mu_k^*)}{f(\mu_k^t)} \ \  \frac{g(\mu_k^t\mid \mu_k^*)}{g(\mu_k^*\mid \mu_k^t)} = \left(\frac{1-\Phi\left(-\frac{\mu_k^t}{\sigma_k}\right)
}{1-\Phi\left(-\frac{\mu_k^*}{\sigma_k}\right)}\right)^{N L_k} \times \exp \left[ \frac{-\sum_{n=1}^N\sum_{l=1}^{L_k} \ (s_{nl}-\mu_k^*)^2}{2\ \sigma_k^2}-\frac{(\mu_k^*-\mu_0)^2}{2\ \sigma_0^2}\right] 
\end{equation*}

\begin{equation*}
\times \ \exp\left[\frac{\sum_{n=1}^N\sum_{l=1}^{L_k} \ (s_{nl}-\mu_k^t)^2}{2\ \sigma_k^2}+\frac{(\mu_k^t-\mu_0)^2}{2\ \sigma_0^2}\right].
\end{equation*}
\\

\subsection{Sampling the parameter $\sigma_k^2$ :}

\begin{equation}
f\left(\sigma_k^2\mid s_{nl},\mu_k,\alpha_0,\beta_0\right) \propto \\ \ \frac{\left(\frac{1}{\sigma_k^2}\right)^{\frac{NL_k}{2}+\alpha_0+1}\ \ \exp \left[\frac{-\sum_{n=1}^N\sum_{l=1}^{L_k} \ (s_{nl}-\mu_k)^2}{2\ \sigma_k^2}-\frac{\beta_0}{\sigma_k^2}\right]}{\left(1-\Phi\left(-\frac{\mu_k}{\sigma_k}\right)\right)^{NL_k}}
\label{eqt:posteriorsigma}
\end{equation}

\text{The proposal law is :}
\begin{center}
\begin{equation*}
g\left(x\mid (\sigma_k^2)^t\right)   \propto \  \ \frac{\frac{1}{\sqrt{2\ \pi}\ \epsilon_4} \ \exp \left[- \frac{(x-(\sigma_k^2)^t)^2}{2\ \epsilon_4^2}\right]}{1-\Phi\left(\frac{-(\sigma_k^2)^t}{\epsilon_4}\right)}. \ \ \ 
\end{equation*}
\end{center}

We generate $\sigma_k^{2*}$, such that : 
\begin{equation*}
 \ \sigma_k^{2*} \sim g\left(x\mid (\sigma_k^2)^t\right).
\end{equation*}
$\textrm{The acceptance-rejection rule is :}$ \\
\begin{equation*}
 \ \ \ \ (\sigma_k^2)^{t+1}=\begin{cases}\sigma_k^{2*} \ \ \textrm{with \ \ prob}\ \  \min \left\{ \frac{f(\sigma_k^{2*})}{f((\sigma_k^2)^t)} \ \  \frac{g((\sigma_k^2)^t\mid \sigma_k^{2*})}{g(\sigma_k^{2*}\mid (\sigma_k^2)^t)} \ ,1\right\}
\\  (\sigma_k^2)^t \ \  \ else\end{cases}
\end{equation*}

with 
\begin{small}
\begin{equation*}
 \frac{f(\sigma_k^{2*})}{f((\sigma_k^2)^t)} \ \  \frac{g((\sigma_k^2)^t\mid \sigma_k^{2*})}{g(\sigma_k^{2*}\mid (\sigma_k^2)^t)} =  \left(\frac{1-\Phi\left(-\frac{\mu_k}{\sigma_k^t}\right)
}{1-\Phi\left(-\frac{\mu_k}{\sqrt{\sigma_k^{2*}}}\right)}\right)^{N L_k} \ \left[\frac{1-\Phi\left(\frac{-(\sigma_k^2)^t}{\epsilon_3}\right)}{1-\Phi\left(\frac{-\sigma_k^{2*}}{\epsilon_3}\right)}\right] \times \ \exp\left[\frac{\sum_{n=1}^N\sum_{l=1}^{L_k} \ (s_{nl}-\mu_k)^2}{2\ (\sigma_k^2)^t}+\frac{\beta_0}{ (\sigma_k^2)^t}\right] 
\end{equation*}
\end{small}

\begin{equation*}
\times \ \left(\frac{(\sigma_k^2)^t}{\sigma_k^{2*}}\right)^{\frac{NL_k}{2}+\alpha_0+1} 
 \ \exp \left[ \frac{-\sum_{n=1}^N\sum_{l=1}^{L_k} \ (s_{nl}-\mu_k)^2}{2\ x^{t}}-\frac{\beta_0}{ \sigma_k^{2*}}\right].  \  
\end{equation*}
\\

\subsection{Sampling the parameter $z_{l}$ :}

\begin{equation}
f\left(z_{l}=k\mid s_{nl},\sigma_k^2,\mu_{k}\right) \propto \ \frac{\frac{1}{(\sqrt{2\ \pi}\ \sigma_k)^N} \ \exp \left[-\sum_{n=1}^N \frac{(s_{nl}-\mu_k)^2}{2\ \sigma_k^2}\right]}{\left[1-\Phi\left(\frac{-\mu_k}{\sigma_k}\right)\right]^N}
\label{eqt:posteriorZ}
\end{equation}

\subsection{Bayesian inference and parameter estimation}
The main steps of the proposed Metropolis-within-Gibbs sampler are summarized in Algo.  \ref{alg:Metropolis}. This algorithm provides a sequence of samples of the vector $\bsX = \left(\bsS,\boldsymbol{\theta},\bsz,\mu_k,\sigma_k^2 \right)$ denoted as $\bsX_{j}^{(i)}$
that are used to
approximate the MMSE estimators by using Monte Carlo integration \cite{Robert1999} as
\begin{equation}
\bsX^{\textrm{MMSE}} \simeq \  \frac{1}{\textrm{N}_{\textrm{MC}}-\textrm{N}_{\textrm{bi}}} \sum_{i=\textrm{N}_{\textrm{bi}}+1}^{\textrm{N}_{\textrm{MC}}}  \bsX^{(i)}
\label{eqt:mmse}
\end{equation}
where  $\textrm{N}_{\textrm{bi}}$ is the number of burn-in iterations and $\textrm{N}_{\textrm{MC}}$ is the total number of Monte Carlo iterations.
 Finally, the following maximum a-posteriori (MAP) estimator is considered for the label $\bsz$

\begin{equation}
\bsz^{\textrm{MAP}}_{l} \simeq \begin{cases}1 \ \ \  \textrm{if} \left[\bsz^{(i)}_{l}=1\right]_{i=\textrm{N}_{\textrm{bi}}+1}^{\textrm{N}_{\textrm{MC}}} \geq \left[\bsz^{(i)}_{l}=2\right]_{i=\textrm{N}_{\textrm{bi}}+1}^{\textrm{N}_{\textrm{MC}}}    \\
2 \ \ \ \textrm{otherwise} \end{cases}
\label{eqt:map}
\end{equation}
where $\left[ x=1 \right]_{i}^{j}$ and $\left[ x=2 \right]_{i}^{j}$ denote the numbers of samples satisfying the conditions $x=1$ and $x=2$ in the interval $[i,j]$.
\subsection{ Convergence:} \label{subsec:Potential} 
Running multiple chains with different initializations allows
to define various convergence measures for MCMC methods \cite{Robertmcmc}. The popular between-within variance criterion has shown
interesting properties for diagnosing convergence of MCMC
methods. This criterion was initially studied by Gelman and
Rubin in \cite{Gelman1992} and has been used in many studies including \cite[p. 33]{Robertmcmc}, \cite{Godsill1998} and \cite{DjuricChun2002}.
The main idea is to run $M$ parallel chains of length $\textrm{N}_\textrm{r}$ + $\textrm{N}_{\textrm{bi}}$ for each data set with different starting
values and to evaluate the dispersion of the estimates obtained
from the different chains. The between-sequence variance $B$ 
and within-sequence variance $W$ for the $M$ Markov chains are defined by 

\begin{equation}
B= \frac{\textrm{N}_\textrm{r}}{M-1} \sum_{m=1}^M \left(\bar{k}_m-\bar{k}\right)^2
\end{equation}
\begin{equation}
W= \frac{1}{M} \sum_{m=1}^M\frac{1}{\textrm{N}_\textrm{r}}\sum_{t=1}^{N_r} \left(k_m^{(t)}-\bar{k}_m\right)^2
\end{equation}
with
\begin{equation}
\bar{k}_m=\frac{1}{\textrm{N}_\textrm{r}}\sum_{t=1}^{\textrm{N}_\textrm{r}} k_m^{(t)}, \ \ \bar{k}= \frac{1}{M}\sum_{m=1}^M\bar{k}_m, \ \ \textrm{N}_\textrm{r}=\textrm{N}_{\textrm{MC}}-\textrm{N}_{\textrm{bi}}. 
\end{equation}
where $k$ is the parameter of interest and $k_m^{(t)}$ is its estimate at the $t$th run of $m$th chain. The convergence of the chain can then
be monitored by the so-called  {\it potential scale reduction factor} $\hat{\rho}$ defined as \cite[p. 332]{Gelman} 
\begin{equation}
\sqrt{\hat{\rho}}=\sqrt{\frac{1}{W} \left(\frac{\textrm{N}_\textrm{r}-1}{\textrm{N}_\textrm{r}} \ W+ \frac{1}{\textrm{N}_\textrm{r}} B\right)}.
\end{equation}


\section{Simulation results} 
\subsection{Synthetic data} \label{subsec:Simulation_results_on_synthetic_data}
\begin{figure}[h!]
\centering
			 \subfloat[\textcolor{red}{}]{\includegraphics[width=1\figwidth, height=10cm]{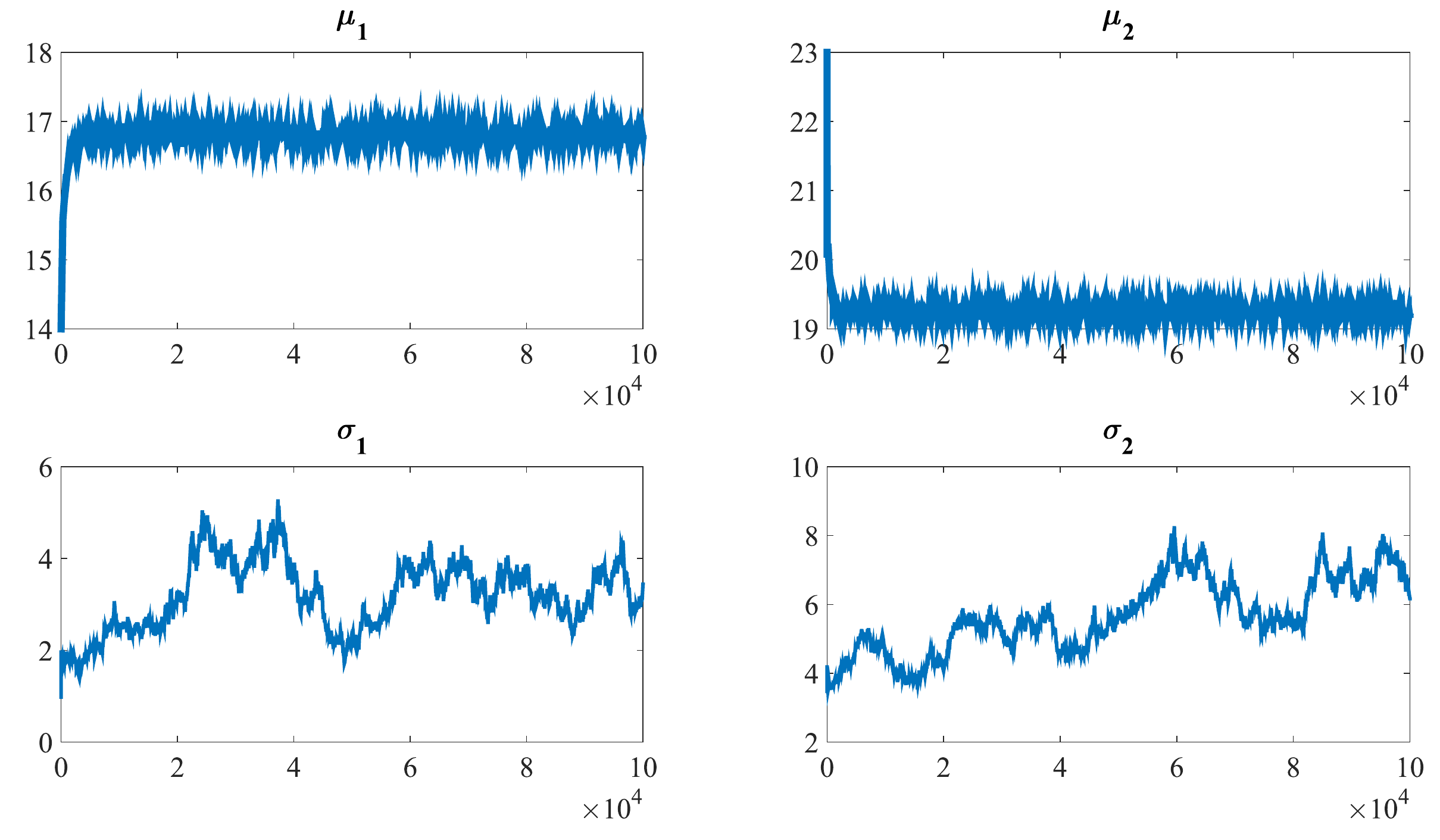}}\par
        \subfloat[\textcolor{red}{}]{\includegraphics[width=1\figwidth, height=9cm]{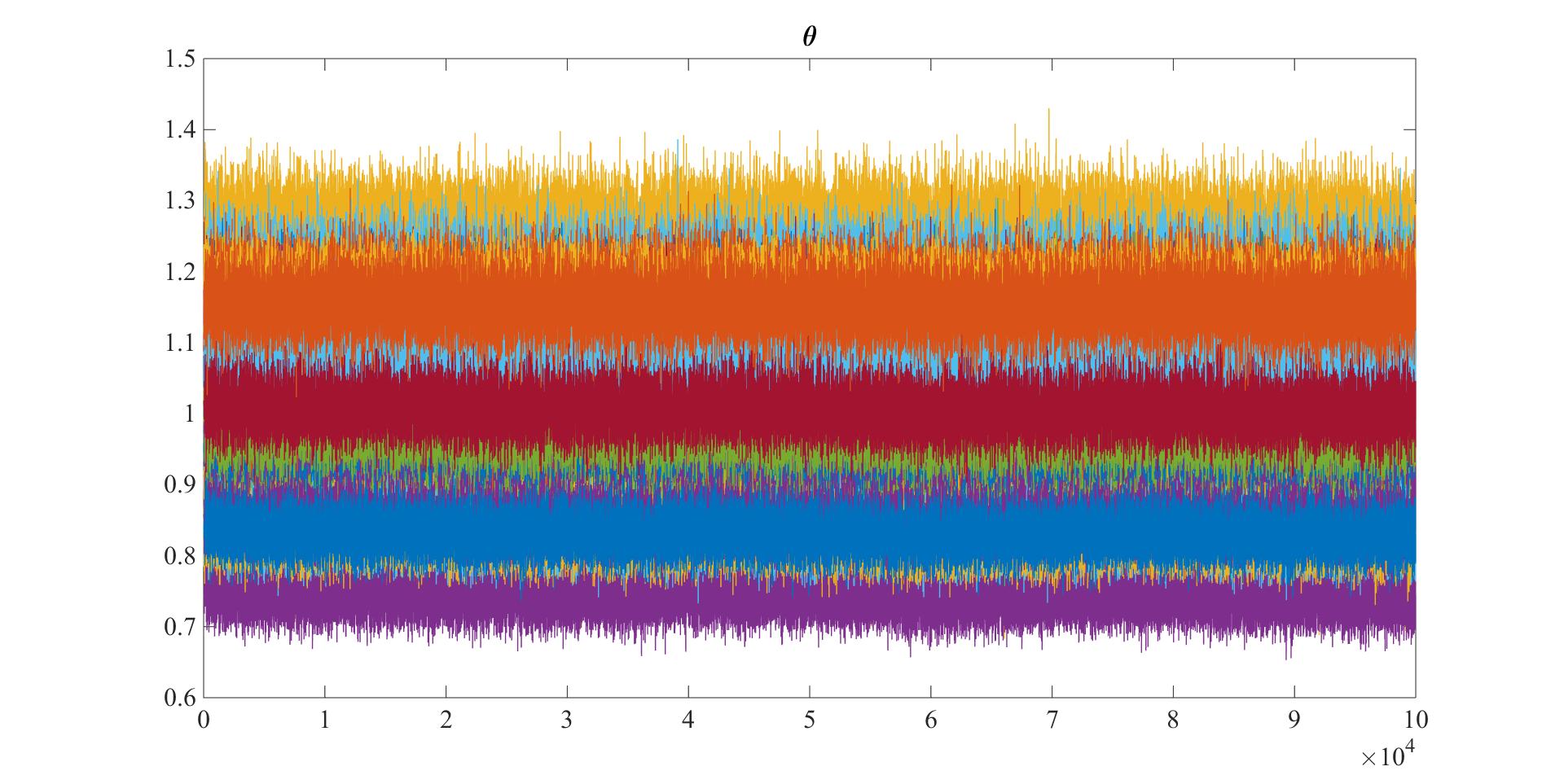}}
				\caption{ Evolution of the convergence of the Markov chains for the different parameters $\hat{\mu_1}, \hat{\mu_2}, \hat{\sigma_1}, \hat{\sigma_2}, \boldsymbol{\theta}$ estimated for the synthetic data with $\textrm{SNR}_Y= 0 \ \textrm{dB}$.}
				\label{fig:chaines_converg_0}
				\end{figure}

\begin{figure}[h!]
\centering
			 \subfloat[\textcolor{red}{}]{\includegraphics[width=1\figwidth, height=10cm]{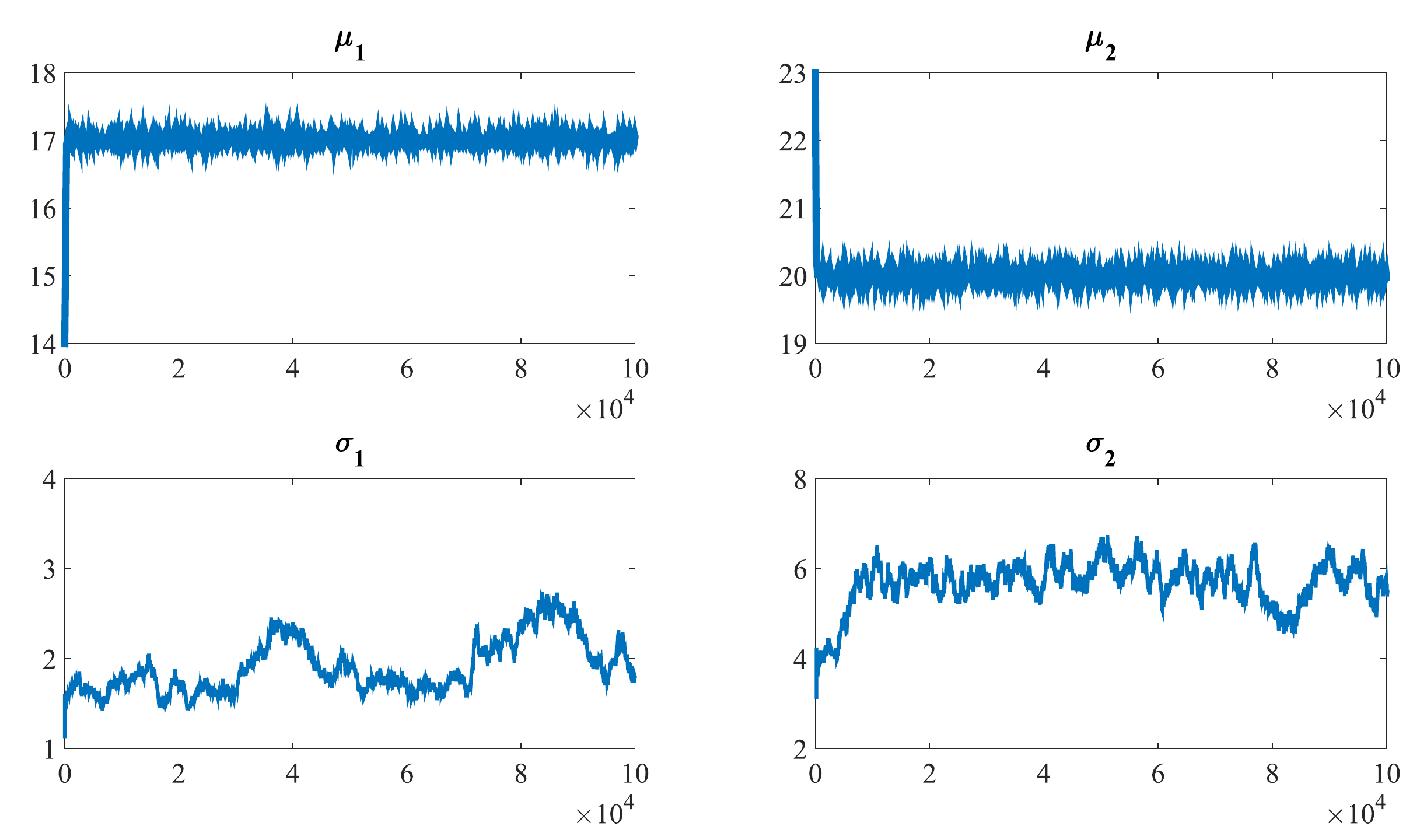}}\par
        \subfloat[\textcolor{red}{}]{\includegraphics[width=1\figwidth, height=9cm]{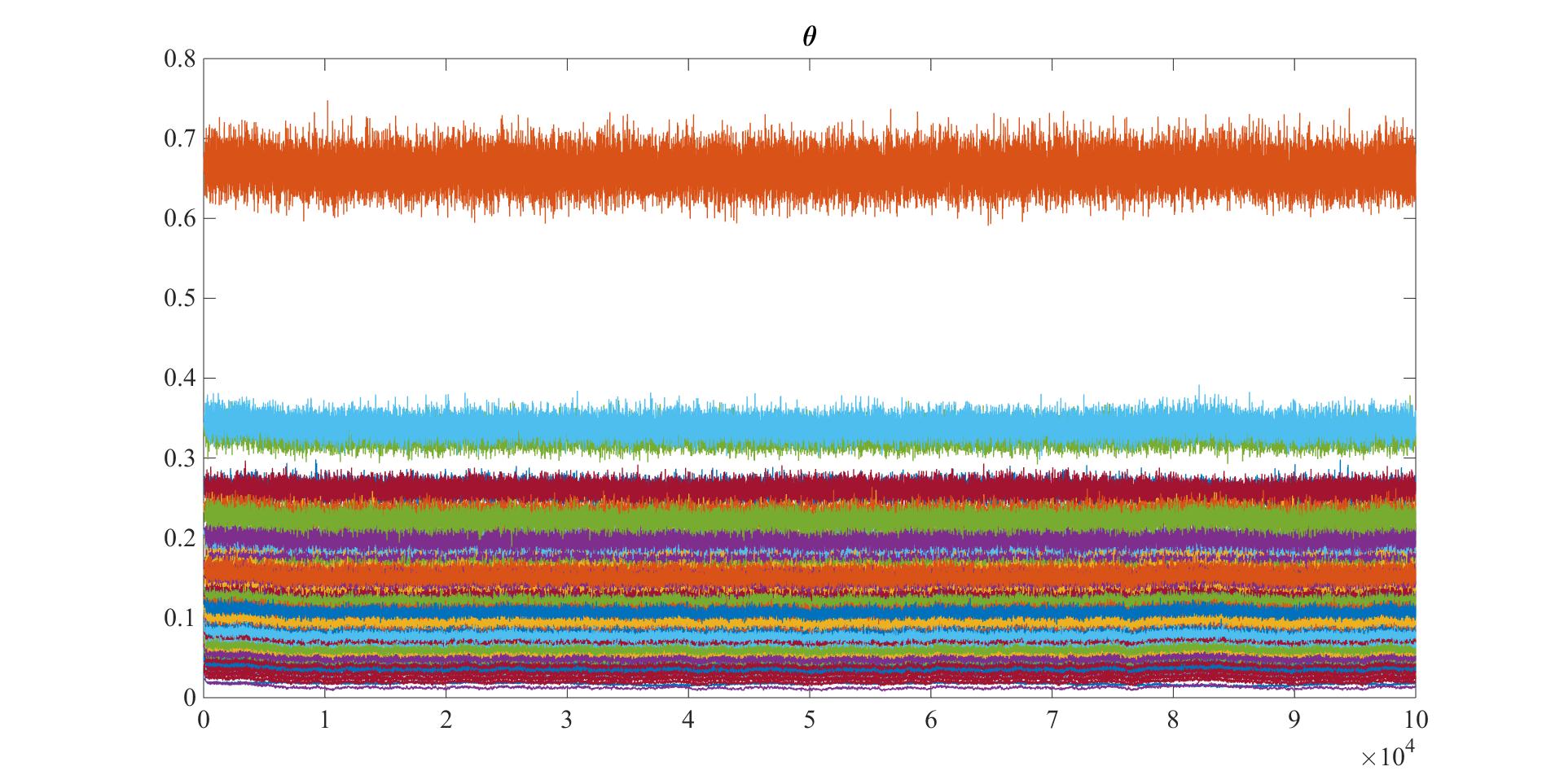}}
				\caption{ Evolution of the convergence of the Markov chains for the different parameters $\hat{\mu_1}, \hat{\mu_2}, \hat{\sigma_1}, \hat{\sigma_2}, \boldsymbol{\theta}$ estimated for the synthetic data with $\textrm{SNR}_Y= 10 \ \textrm{dB}$.}
				\label{fig:chaines_converg_10}
				\end{figure}

\begin{figure}[h!]
\centering
			 \subfloat[\textcolor{red}{}]{\includegraphics[width=1\figwidth, height=10cm]{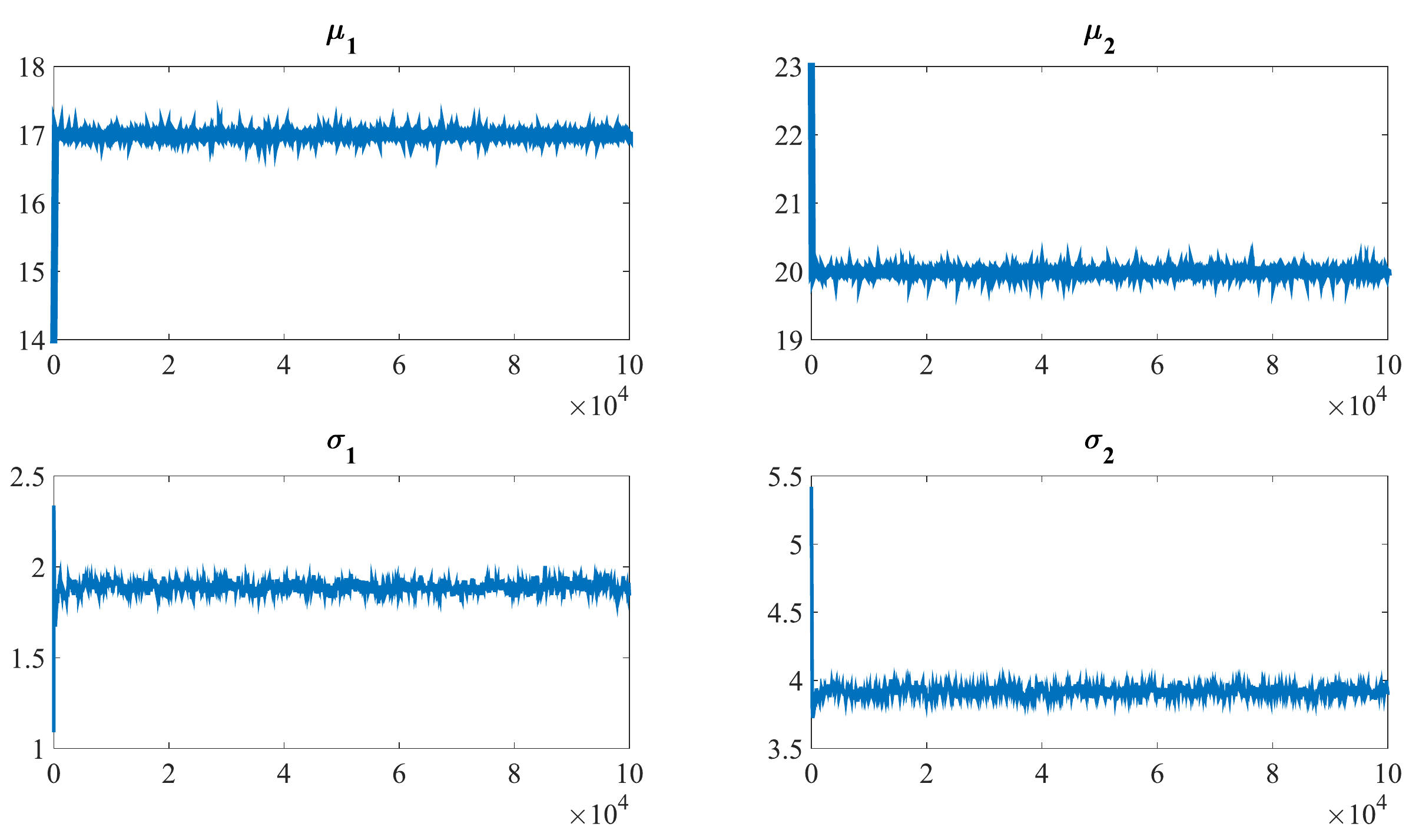}}\par
        \subfloat[\textcolor{red}{}]{\includegraphics[width=1\figwidth, height=9cm]{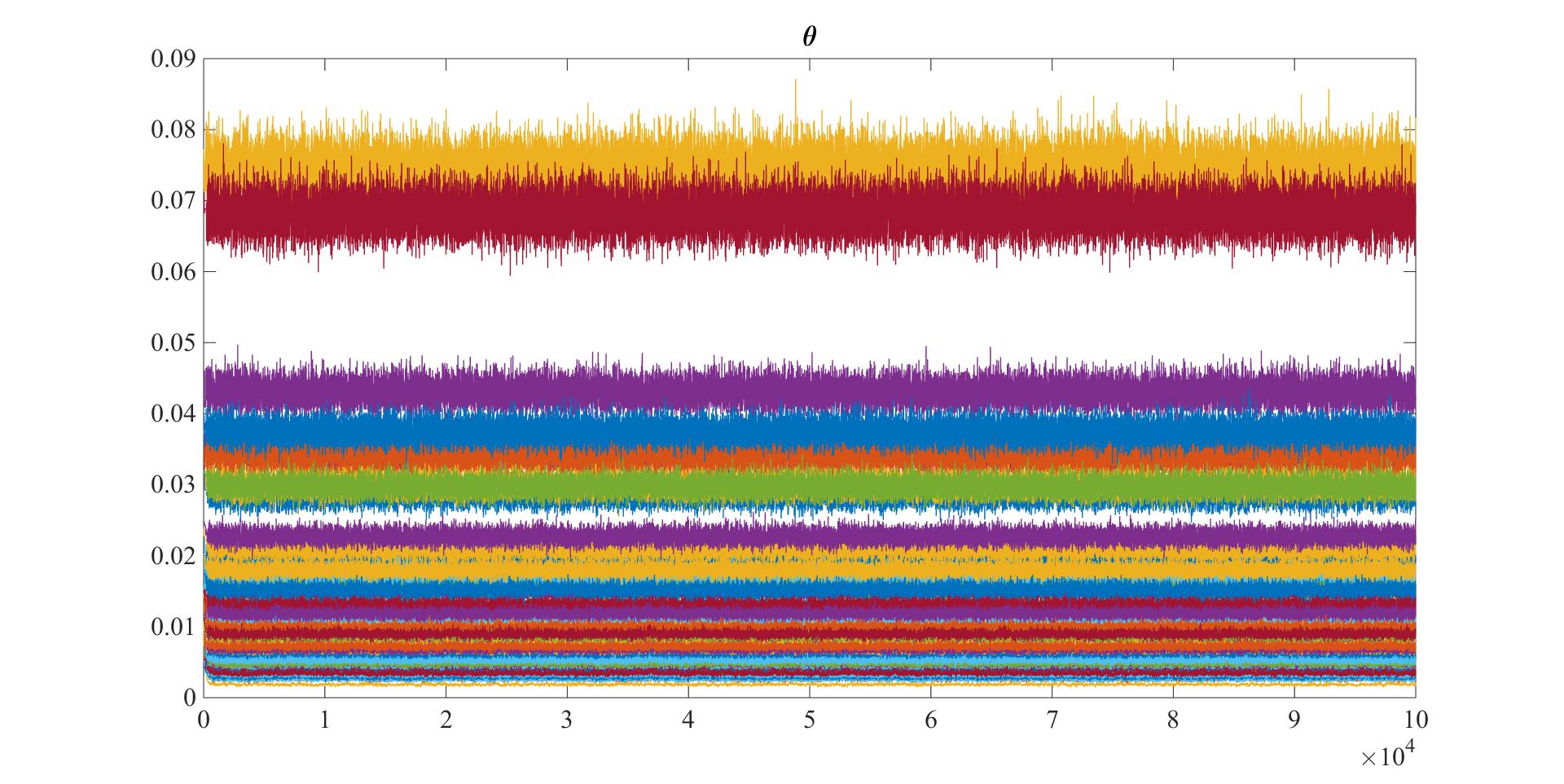}}
				\caption{ Evolution of the convergence of the Markov chains for the different parameters $\hat{\mu_1}, \hat{\mu_2}, \hat{\sigma_1}, \hat{\sigma_2}, \boldsymbol{\theta}$ estimated for the synthetic data with $\textrm{SNR}_Y= 20 \ \textrm{dB}$.}
				\label{fig:chaines_converg_20}
				\end{figure}

\begin{figure}[h!]
\centering
			 \subfloat[\textcolor{red}{}]{\includegraphics[width=1\figwidth, height=10cm]{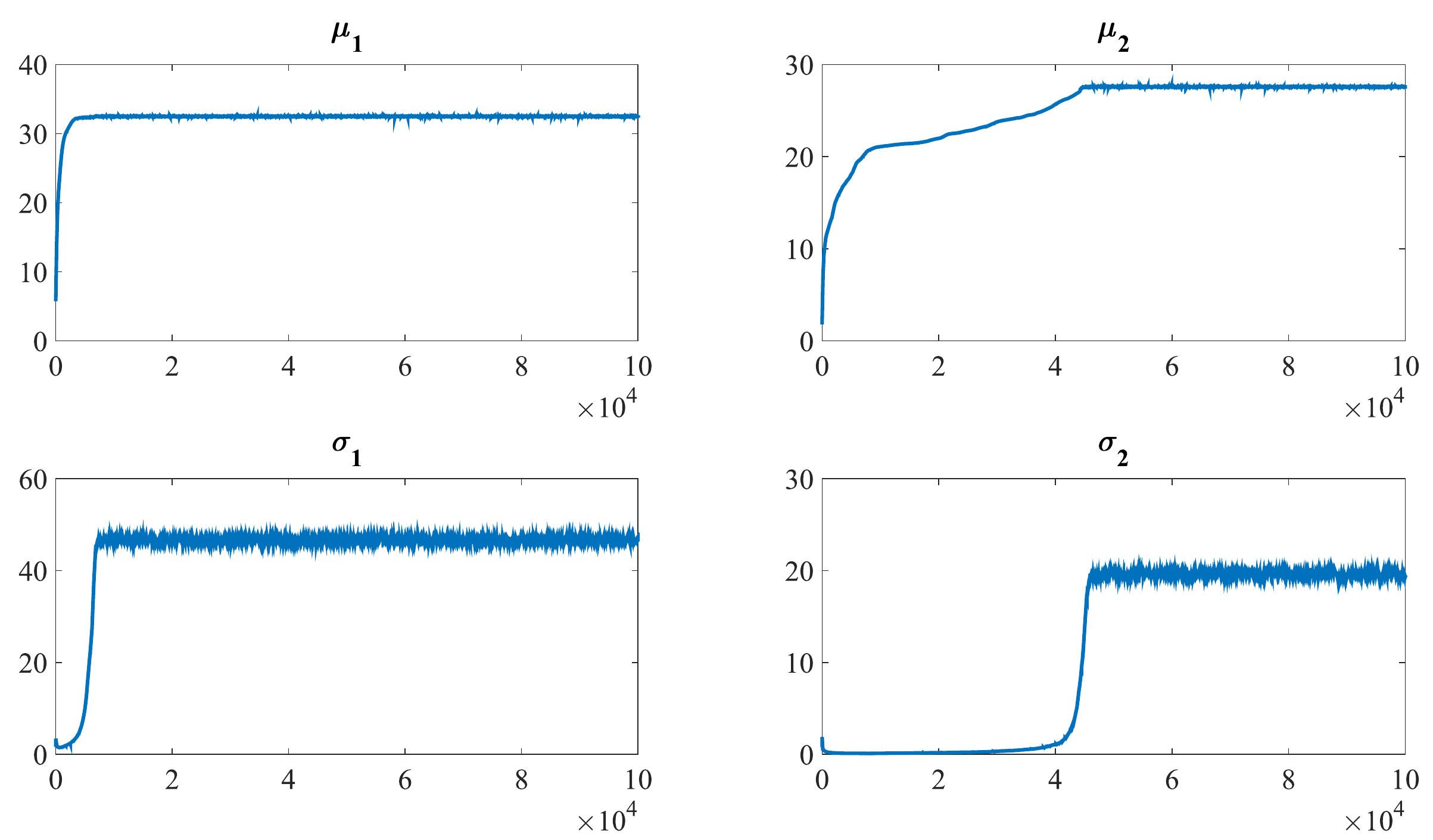}}\par
        \subfloat[\textcolor{red}{}]{\includegraphics[width=1\figwidth, height=9cm]{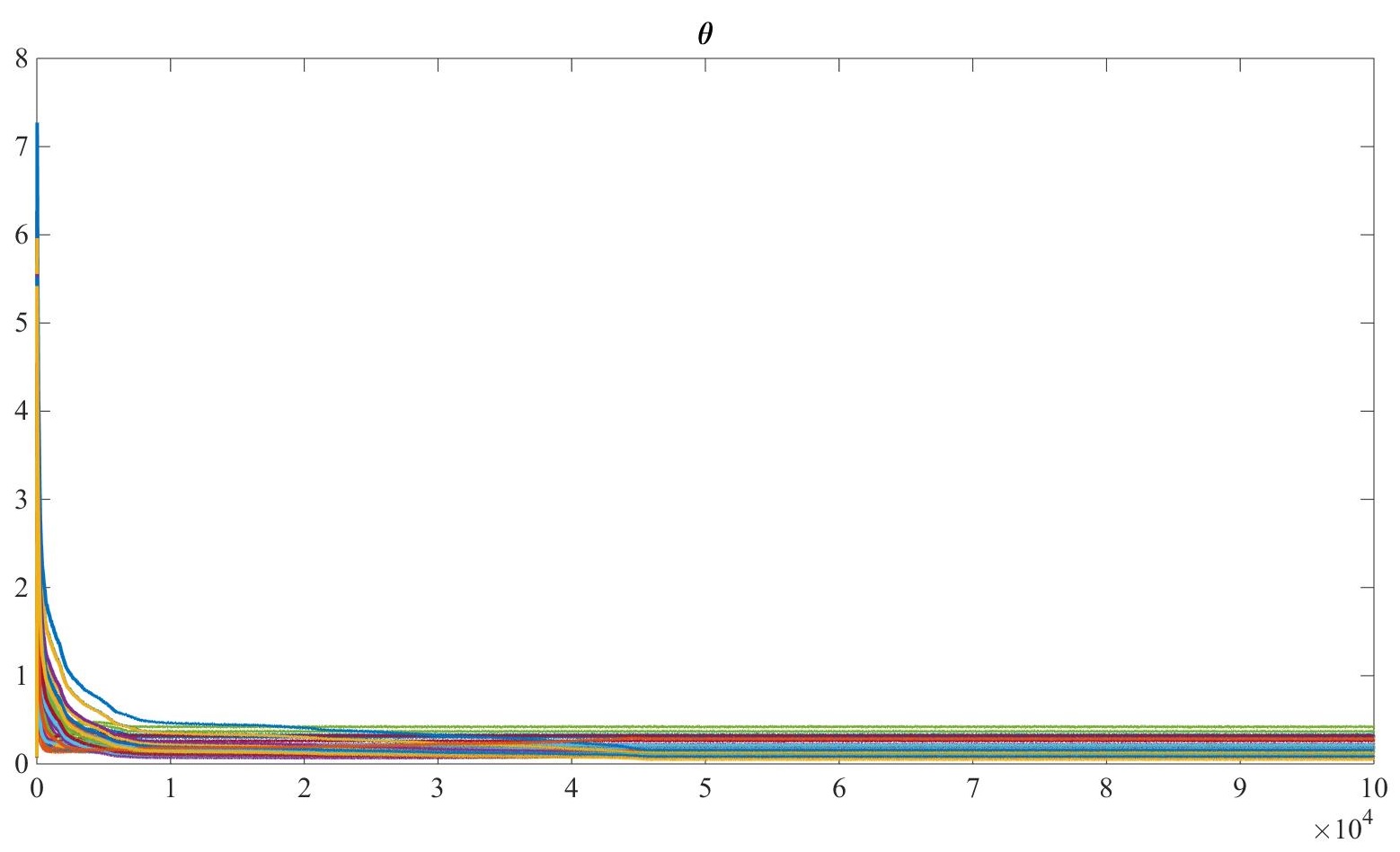}}
				\caption{ Evolution of the convergence of the Markov chains for the different parameters $\hat{\mu_1}, \hat{\mu_2}, \hat{\sigma_1}, \hat{\sigma_2}, \boldsymbol{\theta}$ estimated for the real RCM images.}
				\label{fig:chaines_converg}
				\end{figure}

This section evaluates the performance of the proposed algorithm on synthetic data. Different experiments were conducted using three values of the signal to noise ratio $\textrm{SNR} \in \{ 0 \ \textrm{dB}, 10 \ \textrm{dB}, 20  \ \textrm{dB} \}$, allowing the algorithm performance to be appreciated for different noise levels. This section considers $L = 100$ synthetic
images. Each image contains $N = 2000$ pixels and was generated according to \eqref{eqt:likelihood2}. These images were separated into healthy and lentigo classes containing $50$ images. The noiseless images of the two classes were respectively generated according to the truncated Gaussian distributions $\mathcal{N}_{\mathds{R^+}}(\mu_1,\sigma_1^2)$ and $\mathcal{N}_{\mathds{R^+}}(\mu_2,\sigma_2^2)$, with $\mu_1=17,\mu_2=20, \sigma_1^2=2, \sigma_2^2=4$. The sampler convergence of the algorithm is monitored by computing the potential
scale reduction factor introduced in \eqref{subsec:Potential} for an
appropriate parameter of interest. Different choices for the parameter $k$ could be considered for the proposed method.  This paper proposes to monitor the convergence of the
Metropolis-within-Gibbs sampler by checking the noise variance $\theta$ (see \cite{Dobigeon2008,Godsill1998} for
a similar choice). The potential scale reduction factor for parameter $\theta$ computed for $M =10$ Markov chains is equal
to 1.01. This value of $\sqrt{\hat{\rho}}$ confirms the good convergence of
the sampler (a recommendation for convergence assessment is
a value of $\sqrt{\hat{\rho}} \leq 1.2 $ \cite[p. 332]{Gelman}  ). Figs. \ref{fig:chaines_converg_0}, \ref{fig:chaines_converg_10}, \ref{fig:chaines_converg_20} and \ref{fig:chaines_converg} show the evolution of the Markov chains for the different parameters $\hat{\mu_1}, \hat{\mu_2}, \hat{\sigma_1}, \hat{\sigma_2}, \boldsymbol{\theta}$ estimated for synthetic data with $\textrm{SNR}_Y= [ 0 \ \textrm{dB}, 10 \ \textrm{dB}, 20 \ \textrm{dB}]$ and the real data using the RCM images, respectively.
Algo. \ref{alg:Metropolis} was run for $\textrm{N}_{\textrm{MC}}=100000$ iterations and the different model parameters were estimated using \eqref{eqt:mmse} and \eqref{eqt:map} using a burn-in period of length $\textrm{N}_{\textrm{bi}}=99900$. The performance of the algorithm was evaluated by computing the mean square errors (MSEs) of the different parameters and the signal to noise ratios (SNRs) defined as
  \begin{equation}
  \textrm{MSE}_j=\parallel \hat{\bsX_j}-\bsX_j\parallel^2
  \end{equation}
 \begin{equation}
  \textrm{SNR}_j = 20 \log_{10} \left(\frac{{||\bsX_j||}}{{||\bsX_j- \widehat{\bsX}_j||} } \right).
 \end{equation}
Quantitative results are presented in Table \ref{tab:Performance_on_synth} for the three experiments. This table shows good estimation results of the parameters when considering different noise levels. This table also shows excellent classification results for $\textrm{SNR}_Y\geq 10$ dB, and  $91\%$ when considering the challenging case $\textrm{SNR}_Y= 0$ dB. These results highlight the potential of the proposed strategy in denoising and classifying the images obtained from model \eqref{eqt:likelihood2} and improving the estimation of the different parameters of this model.


\begin{table*}[]
\centering
\caption{Performance of the proposed algorithm for denoising and classification of synthetic data for three  corrupted data $\textrm{SNR}_Y= [ 0 \ \textrm{dB}, 10 \ \textrm{dB}, 20 \ \textrm{dB}].$ }
\label{tab:Performance_on_synth}
{\renewcommand{\arraystretch}{1.3}
\begin{tabular}{c|c|c|c|c|c|c|l}
\cline{2-7}
\multicolumn{1}{l|}{}                         & \multicolumn{2}{c|}{$\textrm{SNR}_Y= 0 \ \textrm{dB}$} & \multicolumn{2}{c|}{$\textrm{SNR}_Y= 10 \ \textrm{dB}$} & \multicolumn{2}{c|}{$\textrm{SNR}_Y= 20 \ \textrm{dB}$} &  \\ \cline{2-7}
\multicolumn{1}{l|}{}                         & $\textrm{MSE}^{2}$        & $\textrm{SNR (dB)}$        & $\textrm{MSE}^{2}$         & $\textrm{SNR (dB)}$        & $\textrm{MSE}^{2}$         & $\textrm{SNR (dB)}$        &  \\ \cline{1-7}
\multicolumn{1}{|c|}{${\mu_1}$}           & 0.56                      & 30.12                      & $1.54.10^{-4}$             & 62.72                      & $2.63.10^{-5}$             & 70.4                       &  \\ \cline{1-7}
\multicolumn{1}{|c|}{${\mu_2}$}           & 0.95                      & 21.42                       & $1.89.10^{-5}$             & 73.24                      & $6.64.10^{-5}$             & 67.79                      &  \\ \cline{1-7}
\multicolumn{1}{|c|}{${\sigma_1^2}$}      & 2.91                      & 1.01                       & 0.015                      & 18.07                       & 0.011                      & 25.7                      &  \\ \cline{1-7}
\multicolumn{1}{|c|}{${\sigma_2^2}$}      & 7.14                      & 2.57                       & 4.58                       & 5.42                       & 0.006                      & 22.07                      &  \\ \cline{1-7}
\multicolumn{1}{|c|}{${\boldsymbol{\theta}}$}          & $1.14.10^{-3}$            & 20.44                       & $4.74.10^{-5}$             & 26.56                      & $5.68.10^{-7}$             & 30.44                    &  \\ \cline{1-7}
\multicolumn{1}{|c|}{${\bsS}$}            & 5.48                      & 16.53                      & 2.88                       & 20.81                      & 0.7093                     & 26.87                      &  \\ \cline{1-7}
\multicolumn{1}{|c|}{Accuracy} & \multicolumn{2}{c|}{91$\%$}                           & \multicolumn{2}{c|}{100$\%$}                            & \multicolumn{2}{c|}{100$\%$}                            &  \\ \cline{1-7}
\multicolumn{1}{|c|}{Accuracy (CART)} & \multicolumn{2}{c|}{83$\%$}                           & \multicolumn{2}{c|}{100$\%$}                            & \multicolumn{2}{c|}{100$\%$}                            &  \\ \cline{1-7}
\end{tabular}}
\end{table*}
\subsection{Real data}
\label{sec:Results_on_Jason2_real_data}
\begin{figure}[h!]
\centering
\includegraphics[width=1\figwidth, height=7cm]{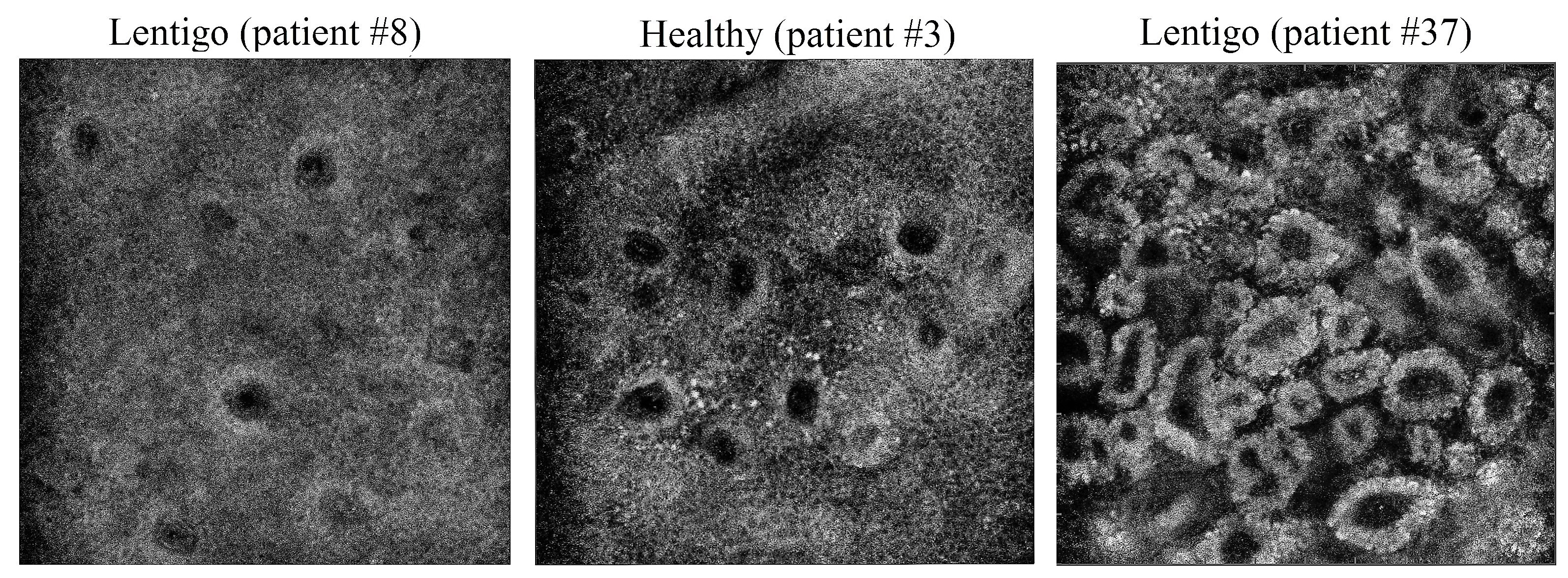}
\caption{ Images (at the depth 49.5 $\mu m$) from the patient $\#8$ who is badly classified compared to a healthy and lentigo patient (well classified). One can observe more similarity between this patient and the healthy one then with the lentigo. } \label{fig:images}
\end{figure}

\begin{figure}[h!]
\centering
\includegraphics[width=1\figwidth, height=11cm]{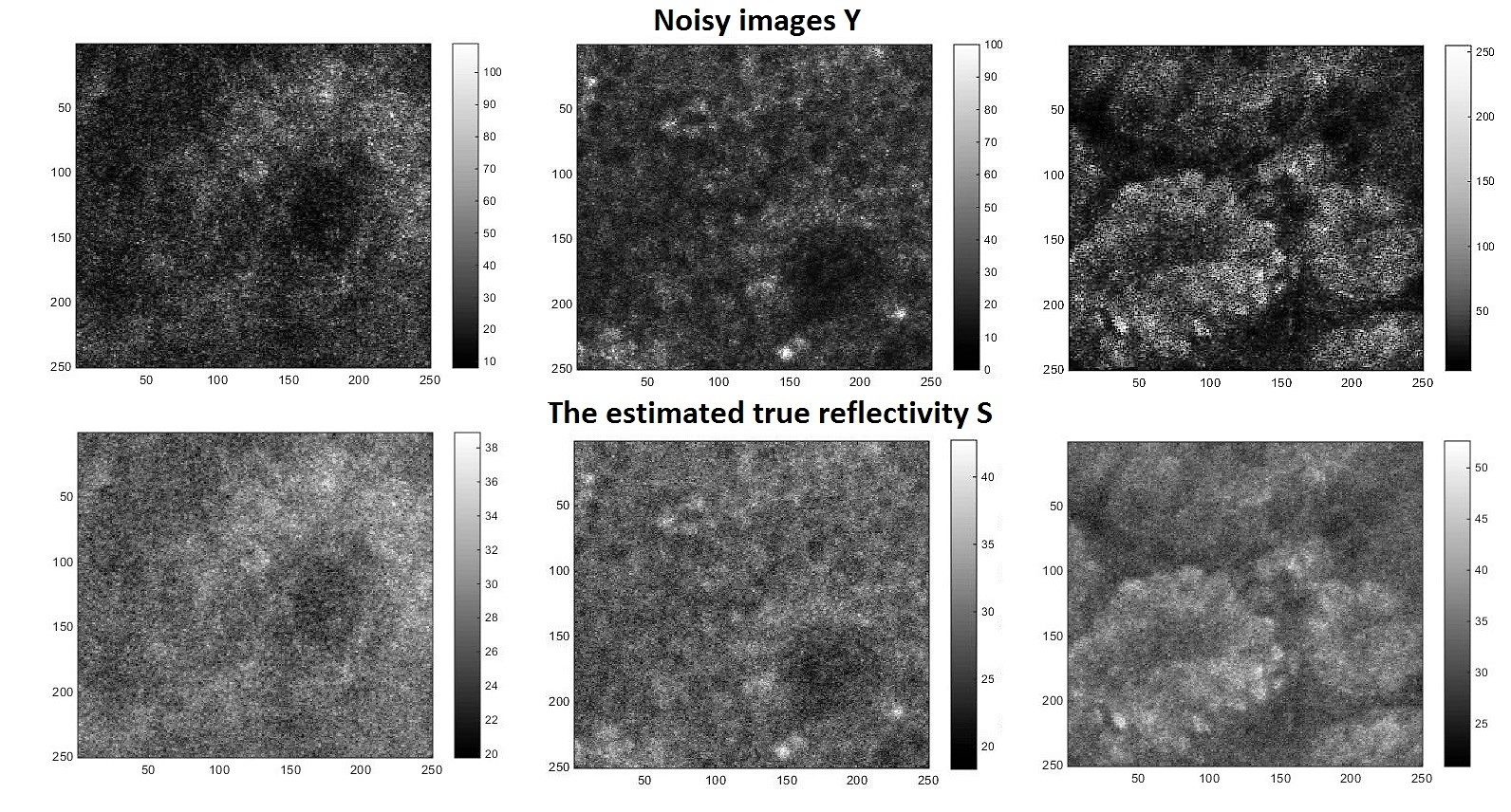}
\caption{ Examples of noisy images (at the depth 49.5 $\mu m$) and their estimated true reflectivities. } \label{fig:images_debruite}
\end{figure}

This section is devoted to the validation of the proposed denoising and classification algorithm when applied to real RCM images. These RCM images were acquired with apparatus Vivascope $1500$ and correspond to the stratum corneum, the epidermis layer, the dermis-epidermis junction (DEJ) and the upper papillary dermis. Each RCM image shows a $500 \times 500 \mu m$ field of view with $1000 \times 1000$ pixels.  A set of $L=45$ women aged $60$ years and over were recruited.  All the volunteers gave their informed consent for examination of skin by RCM. According to the clinical evaluation performed by a physician, volunteers were divided into two groups.  The first group was formed by $27$ women with at least $3$ lentigines on the back of the hand whereas $18$ women without lentigo  constituted the control group. Images were taken on lentigo lesions for volunteers of the first group and on healthy skin on the back of the hand for the control group. Consequently, our database contained $M$ = 45 patients. An examination of each acquisition was performed in order to locate the stratum corneum and the DEJ precisely  in each image.  Given the large size of the images, we preferred to select and apply our algorithm to patches of $250 \times 250$ pixels for each image to reduce the computational cost. The obtained results  were then used to calculate the confusion matrix and four indicators (sensitivity, specificity, precision, accuracy) shown in Tables \ref{tab:Performance_on_real} and \ref{tab:Performance_on_real_cart}.  These indicators are defined as Sensitivity = TP/(TP+FN), Specificity = TN/(FP+TN), Precision = TP/(TP+FP), Accuracy = (TP+TN)/(TP+FN+FP+TN), where TP, TN, FP and FN are the numbers of true positives, true negatives, false positives and false negatives. This table allows us to  evaluate the classification performance  of the proposed strategy.  The accuracy of the proposed method equals $97.7 \%$, which corresponds to a single mistake  for the lentigo patient $\#8$. Fig. \ref{fig:images} shows that the texture of  this mis-classified image is not very destructed as  for other lentigo patients, and is visually similar to the texture of healthy patients. Fig. \ref{fig:images_debruite} shows examples of noisy RCM images and their estimated true reflectivity, illustrating the denoising part of the proposed algorithm. We can observe that the estimated images have low intensities compared to the noisy images which is due to the fact that the noise is  multiplicative. To assess the significance of our results, our algorithm was then compared to the method presented in \cite{koller2011vivo}. This method consists in extracting from each RCM image a set of 39 analysis parameters (further technical details are available in \cite{wiltgen2008automatic}) and to apply to these features a classification procedure based on classification and regression trees (CART). Note that the CART algorithm was tested on the real RCM images using a leave one out procedure. As shown in Table \ref{tab:Performance_on_real_cart}, the accuracy obtained with the CART algorithm is $82.2 \%$ , i.e., it is slightly smaller that the one obtained with the proposed method. Moreover, the proposed Bayesian model can be used for the characterization of RCM images thanks to its estimated parameters.
 
\begin{table}[]
\centering
\caption{Classification performance on real data (45 patients) using the proposed method.}
\label{tab:Performance_on_real}
{\renewcommand{\arraystretch}{1.5}
\begin{tabular}{|c|c|c|c}
\hline
\textbf{Confusion matrix} &$\widehat{\textbf{L}}$ & $\widehat{\textbf{H}}$ & \multicolumn{1}{c|}{\textbf{\begin{tabular}[c]{@{}c@{}}Sensitivity\\ Specificity\end{tabular}}} \\ \hline
\textbf{Lentigo}          & \textbf{26  }             & \textbf{1                } & \multicolumn{1}{c|}{\textbf{96.2  $\boldsymbol\%$ }}                                                                  \\ \hline
\textbf{Healthy}          & \textbf{0 }                 & \textbf{18               } & \multicolumn{1}{c|}{\textbf{100 $\boldsymbol\%$ }}                                                                   \\ \hline
\textbf{Precision}        & \textbf{100 $\boldsymbol\%$ \ }           & \textbf{94.7 $\boldsymbol\%$  }        & \textbf{}                                                                                       \\ \cline{1-3}
\textbf{Accuracy}         & \multicolumn{2}{c|}{\textbf{97.7 $\boldsymbol\%$ }}          & \textbf{}                                                                                       \\ \cline{1-3}
\end{tabular}}
\end{table}

\begin{table}[]
\centering
\caption{Classification performance on real data (45 patients) using the CART method.}
\label{tab:Performance_on_real_cart}
{\renewcommand{\arraystretch}{1.5}
\begin{tabular}{|c|c|c|c}
\hline
\textbf{Confusion matrix} &$\widehat{\textbf{L}}$ & $\widehat{\textbf{H}}$ & \multicolumn{1}{c|}{\textbf{\begin{tabular}[c]{@{}c@{}}Sensitivity\\ Specificity\end{tabular}}} \\ \hline
\textbf{Lentigo}          & \textbf{24}             & \textbf{3} & \multicolumn{1}{c|}{\textbf{88.8 $\boldsymbol\%$}}                                                                  \\ \hline
\textbf{Healthy}          & \textbf{5}                 & \textbf{13} & \multicolumn{1}{c|}{\textbf{72.2 $\boldsymbol\%$}}                                                                   \\ \hline
\textbf{Precision}        & \textbf{82.7 $\boldsymbol\%$}           & \textbf{81.2 $\boldsymbol\%$ }        & \textbf{}                                                                                       \\ \cline{1-3}
\textbf{Accuracy}         & \multicolumn{2}{c|}{\textbf{82.2 $\boldsymbol\%$}}          & \textbf{}                                                                                       \\ \cline{1-3}
\end{tabular}}
\end{table}  

\section{Conclusions} \label{sec:Conclusions}
This paper presented a new Bayesian strategy as well as an MCMC
algorithm for classifying RCM images as healthy or lentigo images. A Bayesian model was introduced based on a gamma distribution for the multiplicative speckle noise and on various priors assigned to the unknown model parameters. A hybrid Gibbs sampler was then considered to sample the posterior of this Bayesian model and to build Bayesian estimators.  Simulation results conducted on synthetic and real data allowed the good performance of the proposed classifier to be appreciated. Future work includes the introduction of spatial correlation on the estimated noiseless images to improve their quality.

\bibliographystyle{IEEEtran}
\bibliography{biblio_all}
\end{document}

%% file: notations.tex
\input com_notations.tex

\newcounter{algo}
\renewcommand{\thealgo}{\arabic{algo}}

%% file: com_notations.tex
\def\bsb{{\boldsymbol{b}}}

\def\bss{{\boldsymbol{s}}}

\def\bsy{{\boldsymbol{y}}}
\def\bsz{{\boldsymbol{z}}}

\def\bsS{{\boldsymbol{S}}}

\def\bsX{{\boldsymbol{X}}}
\def\bsY{{\boldsymbol{Y}}}

